\documentclass[conference]{IEEEtran}

\usepackage{hyperref}
\hypersetup{
	 colorlinks   = true,
     citecolor    = black,
     linkcolor    = black,
     urlcolor     = black
}
\usepackage{graphicx}                       
\usepackage{stfloats}
\usepackage[tight,footnotesize]{subfigure}
\usepackage{amssymb,amsmath}
\usepackage{textcomp}
\usepackage{mdwmath}
\usepackage{mdwtab}
\usepackage{eqparbox}
\usepackage{rotating}                     
\usepackage{array}                        
\usepackage{listings}                     
\usepackage{algorithm}                    
\usepackage[utf8]{inputenc}
\usepackage[noend]{algpseudocode}

\bibliography{references}

\begin{document}
\title{Comparing Image Segmentation Algorithms}

\author{ \IEEEauthorblockN{Milind Cherukuri} \IEEEauthorblockA{ \textit{University of North Texas} \\ Texas, USA \\ Email: \texttt{cherukurimilind@gmail.com} } }
\maketitle

\begin{abstract}

This paper presents a novel approach for denoising binary images using simulated annealing (SA), a global optimization technique that addresses the inherent challenges of non-convex energy functions. Binary images are often corrupted by noise, necessitating effective restoration methods. We propose an energy function \(E(x, y)\) that captures the relationship between the noisy image \(y\) and the desired clean image \(x\). Our algorithm combines simulated annealing with a localized optimization strategy to efficiently navigate the solution space, minimizing the energy function while maintaining computational efficiency. We evaluate the performance of the proposed method against traditional iterative conditional modes (ICM), employing a binary image with 10\% pixel corruption as a test case. Experimental results demonstrate that the simulated annealing method achieves a significant restoration improvement, yielding a 99.19\% agreement with the original image compared to 96.21\% for ICM. Visual assessments reveal that simulated annealing effectively removes noise while preserving structural details, making it a promising approach for binary image denoising. This work contributes to the field of image processing by highlighting the advantages of incorporating global optimization techniques in restoration tasks.

\end{abstract}

\section{Introduction}
\vspace{1em}
Binary images, characterized by two distinct pixel values, are widely utilized in various applications, including document analysis, object recognition, and medical imaging \cite{abu2013skeletonization}. However, these images often suffer from noise due to various factors such as sensor imperfections, environmental conditions, and transmission errors. The presence of noise can significantly degrade the quality of binary images, leading to misinterpretation and reduced performance in subsequent image processing tasks \cite{kanwal2022devil}.

Denoising is an essential pre-processing step that aims to restore the original structure of binary images while effectively eliminating noise \cite{fan2019brief}. Traditional approaches, such as median filtering and morphological operations, often struggle to maintain edge integrity and can introduce artifacts in uniform regions. More sophisticated techniques, like iterative conditional modes (ICM) \cite{haris1996image}, provide improved results by leveraging neighborhood relationships but may become trapped in local optima due to their reliance on greedy optimization.

In this paper, we propose a novel approach for denoising binary images that integrates simulated annealing (SA) with a localized optimization strategy. Simulated annealing \cite{nikolaev2010simulated} is a global optimization technique inspired by the annealing process in metallurgy, which allows for the exploration of the solution space while avoiding local minima \cite{venkateswaran2022application}. By incorporating randomness into the search strategy, SA increases the probability of reaching a global optimum. It is particularly well-suited for the non-convex nature of the energy functions associated with binary image denoising.

We formulate an energy function that quantifies the relationship between the noisy image and the desired clean image, enabling the effective application of simulated annealing to minimize energy. Our experimental results demonstrate that the proposed method significantly outperforms traditional ICM techniques in terms of restoration quality, achieving a higher percentage of pixel agreement with the original image. This work contributes to the field of image processing by showcasing the effectiveness of global optimization techniques in enhancing the quality of binary images in the presence of noise \cite{omran2006particle}.
\vspace{-1em}

\section{Problem Description}
\vspace{0.8em}
In image processing, the quality of binary images is often compromised due to the introduction of noise, which can arise from various sources, including sensor inaccuracies, transmission errors, and environmental conditions \cite{huang1993image}. This paper focuses on the problem of denoising binary images that have been subjected to random noise.

We denote the observed noisy image as an array of binary pixel values \(y_i \in \{-1, +1\}\), where \(i = 1, \ldots, D\) indexes all pixels in the image. The noise-free version of the image, represented by binary pixel values \(x_i \in \{-1, +1\}\), is assumed to be obtained from the original clean image by randomly flipping the sign of a subset of pixels with a certain probability, specifically 10\%. This stochastic process leads to the degradation of the image, complicating subsequent analyses and interpretations.

The primary objective of this research is to accurately recover the original noise-free image \(x\) from the observed noisy image \(y\). To achieve this, we employ a denoising strategy that combines simulated annealing with localized optimization techniques. By leveraging the properties of global optimization, we aim to minimize the energy function that quantifies the discrepancy between the noisy and denoised images, thereby facilitating the restoration of the original image structure and enhancing the overall image quality.

\section{Modeling}
\vspace{1em}
Realizing that the thickness of commotion is little, there ought to be areas of strength for an among $x_i$ and $y_i$. One more earlier information is that adjoining pixels $x_i$ and $x_j$ in a picture are emphatically related. This information suggests that $\{\mathbf{x}, \mathbf{y}\}$ has the Markov property and can be depicted with an undirected chart \cite{bishop2006pattern}, so we can utilize the Markov arbitrary field to show this issue.

There are two sorts of coteries in this chart. The first $\{x_i, y_i\}$ has a related energy capability that communicates the relationship between's these factors. Here we can involve a basic energy capability for them: $-\eta x_i y_i$, where $\eta$ is a positive steady. This will create a lower energy (subsequently reassuring a higher likelihood) when $x_i$ and $y_i$ have a similar sign and a higher energy when they have the contrary sign.

The other sort of coteries is $\{x_i, x_j\}$ where $i$ and $j$ are lists of adjoining pixels. Once more, we maintain that the energy should be lower when the pixels have a similar sign than when they have the contrary sign, thus we pick an energy given by $-\beta x_i x_j$ where $\beta$ is another positive consistent.

Since a potential capability is an erratic, non-negative capability over a maximal coterie, we can increase it by any nonnegative elements of subsets of the inner circle, or identically we can add the comparing energies. In this model, this permits us to add an additional term $hx_i$ for every pixel $i$ in the commotion free picture. This would predisposition the model towards pixel esteems that have one specific sign in inclination to the next. Thus the total energy capability for this model is:
$$
E(\mathbf{x}, \mathbf{y}) = h \sum_{i}x_i - \beta\sum_{\{i, j\}}x_ix_j - \eta\sum_{i}x_iy_i
$$

which defines a joint distribution over x and y given by:

$$
p(\mathbf{x}, \mathbf{y}) = \frac{1}{Z}\exp\{-E(\mathbf{x}, \mathbf{y})\}
$$

where $Z$ is the \textit{partition function} \cite{bishop2006pattern}. Our goal is then defined as finding an $\mathbf{x}$ such that:

$$
\mathbf{x} = {\operatorname{arg\,min}} E(\mathbf{x}, \mathbf{y})
$$

\section{Algorithm and Implementation}
\vspace{1em}
\subsection{Iterated Conditional Modes}
We currently fix the components of $y$ to the noticed qualities given by the pixels of the uproarious picture, which verifiably characterizes a restrictive dissemination $p(x|y)$ over commotion free pictures. This is an illustration of the \textit{Ising model}, which has been broadly concentrated on in measurable physical science.

A direct plan is pick whichever state prompts a lower energy at each step of the hunt. This voracious methodology, depicted in Calculation ~\ref{alg:icm}, is known as the \textit{iterated restrictive modes}, or ICM. Regardless of being a straightforward and powerful system, ICM is inclined to neighborhood optima. Hence, we really want one more methodology for an improved outcome.

\begin{algorithm}
\centering
\caption{Binary image denoising with iterated conditional modes}
\label{alg:icm}
  \begin{algorithmic}[1]
    \Function{Deniose}{$\mathbf{y}$, $\beta$, $\eta$, $h$}
        \Comment{$\mathbf{y}$ is the noisy image}
        \State Initialize $\mathbf{x}$ with $\mathbf{x} = \mathbf{y}$
        \State Initialize $E_{best}$ with $E_{best} = E(\mathbf{x}, \mathbf{y})$
	    \For{$k = 1 \to k_{max}$}
	    	\For{each pixel $x_{i}$}
	    		\State $E_1 = E(\mathbf{x}, \mathbf{y})$
	    		\State $x_{i} = - x_{i}$ \Comment{flip the pixel}
	    		\State $E_2 = E(\mathbf{x}, \mathbf{y})$
	    		\If{$E1 > E2$}
		    		\If{$E_2 < E_{best}$}
		    			\State Record the best energy $E_{best} = E_2$
		    		\EndIf
		    	\Else
		    		\State $x_{i} = - x_{i}$ \Comment{flip the pixel back}
		    	\EndIf
	    	\EndFor
	    \EndFor
      \Return $\mathbf{x}$
    \EndFunction
  \end{algorithmic}
\end{algorithm}

\subsection{Simulated Annealing}
It is easy to see that $E(x,y)$ is a non-convex function of $x$, which implies that there will be multiple local minima for $E$ depending on the initial state. To search for a global optimization strategy, In this project, we use \textit{simulated annealing}(SA), which can be easily integrated with the method of gradient descent. By adding randomness to our searching strategy, we increase the probability of reaching a global optimum (and SA does approximate one in this model. The complete algorithm is described in Algorithm ~\ref{alg:sa}.

\begin{algorithm}
\centering
\caption{Binary image denoising with simulated annealing}
\label{alg:sa}
  \begin{algorithmic}[1]
    \Function{Deniose}{$\mathbf{y}$, $\beta$, $\eta$, $h$}
        \Comment{$\mathbf{y}$ is the noisy image}
        \State Initialize $\mathbf{x}$ with $\mathbf{x} = \mathbf{y}$
        \State Initialize $Ebest$ with $Ebest = E(\mathbf{x}, \mathbf{y})$
	    \For{$k = 1 \to k_{max}$}
	    	\State Compute the temperature $t$ = temperature($k,k_{max}$)
	    	\For{each pixel $x_{i}$}
	    		\State $E_1 = E(\mathbf{x}, \mathbf{y})$
	    		\State $x_{i} = - x_{i}$ \Comment{flip the pixel}
	    		\State $E_2 = E(\mathbf{x}, \mathbf{y})$
	    		\State Compute the transition probability $p = prob(E_1, E_2, t)$
	    		\If{$p > q$ where $q$ is a random number in $[0, 1]$}
		    		\If{$E_2 < E_{best}$}
		    			\State Record the best energy $Ebest = E_2$
		    		\EndIf
		    	\Else
		    		\State $x_{i} = - x_{i}$ \Comment{flip the pixel back}
		    	\EndIf
	    	\EndFor
	    \EndFor
      \Return $\mathbf{x}$
    \EndFunction
  \end{algorithmic}
\end{algorithm}

\begin{description}
\item Remark 1.\hfill \\
The temperature function is a decreasing function of iterations. It must converge to $0$ as $k \to k_{max}$. For this implementation, we use

$$
\text{temperature}(k, k_{max}) = \frac{1}{500}(\frac{1}{k} - \frac{1}{k_{max}})
$$
\item Remark 2.\hfill \\
The probability transition function used for this implementation is
$$
prob(E_1, E_2, t) =  \left\{
     \begin{array}{lr}
       1, & E_1 > E_2 \\
       \mathbf{e}^{\frac{E_1 - E2}{t}}, & E_1 \leq E_2
     \end{array}
   \right.
$$
\end{description}

\subsection{Local Optimization}
With either ICM or SA, we only need to alter the $E$ for values relevant to the flipped pixel at each step. Let $x_i$ be the pixel to be flipped; we can just update $E$ with the new $hx_i + \eta x_i y_i + \sum \beta x_i x_j$ where $x_j$ are the neighboring pixels of $x_i$. Then, we can localize the calculation of $E_1$ and $E_2$, making the implementation much faster.

\section{Experiment Result}
We first take a binary image(for simplicity, we choose a black-and-white one), then flip its pixels with 10\% probability to generate the noisy image. After that, we try to restore the original image by running both ICM and SA denoising and comparing their results. For the energy function, we choose $\beta = 1 \times 10^{-4}, \eta = 2.1 \times 10^{-4}, h = 0$. The maximum number of iteration $k_{max}$ is $30$ for both ICM and SA.

\begin{figure}[ht]
\centering
	\begin{minipage}[b]{.7\linewidth}
		\centering
		\includegraphics[width=1\linewidth]{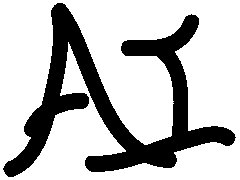}
		\caption{Original image}
		\label{fig:original}
	\end{minipage}\\
	\begin{minipage}[b]{.7\linewidth}
		\centering
		\includegraphics[width=1\linewidth]{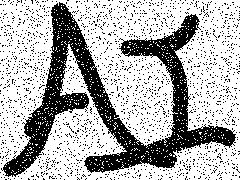}
		\caption{Noisy image with 10\% pixels flipped}
		\label{fig:noisy}
	\end{minipage}
\end{figure}

\begin{figure}[ht]
	\centering
	\begin{minipage}[b]{0.7\linewidth}
		\centering
		\includegraphics[width=\linewidth]{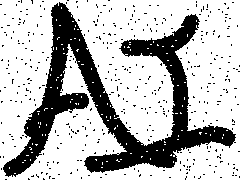}
		\caption{Denoised with ICM}
		\label{fig:icm}
	\end{minipage}\\
	\begin{minipage}[b]{0.7\linewidth}
		\centering
		\includegraphics[width=\linewidth]{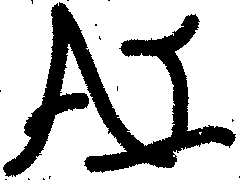}
		\caption{Denoised with SA}
		\label{fig:sa}
	\end{minipage}
\end{figure}

The experiment results are shown in Figure ~\ref{fig:original} - \ref{fig:sa}.


\begin{figure}[ht]
\centering
\includegraphics[width=1\linewidth]{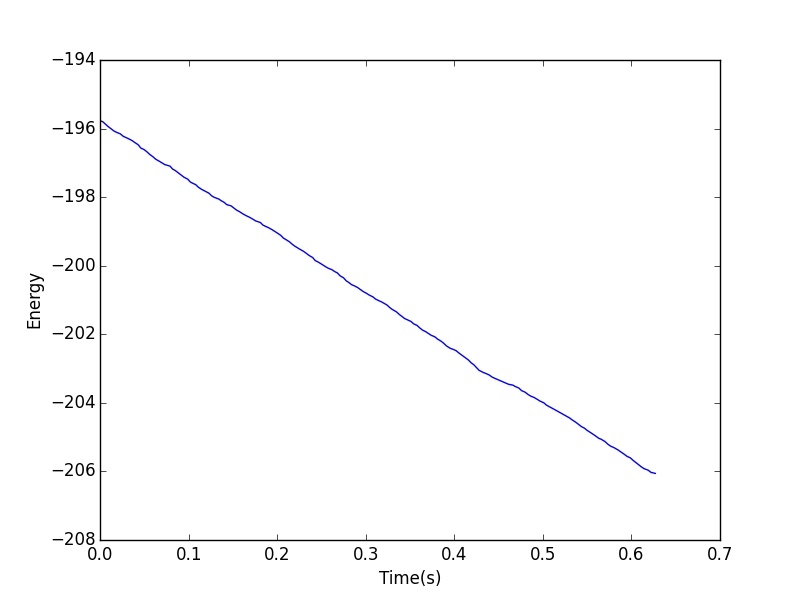}
\caption{Time-Energy series of ICM}
\end{figure}

\begin{figure}[ht]
\centering
\includegraphics[width=1\linewidth]{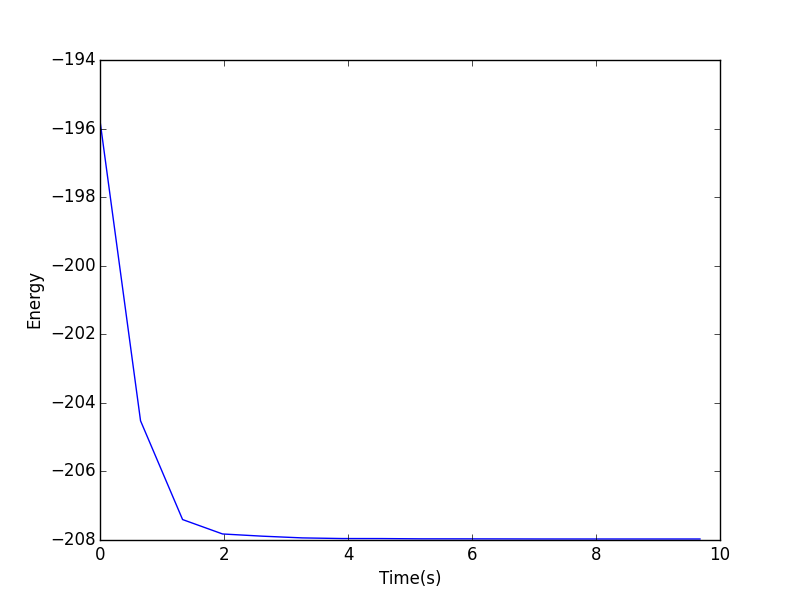}
\caption{Time-Energy series of SA}
\end{figure}
To evaluate the denoised results, we count the non-zero elements in $\mathbf{x} - \mathbf{y}$. ICM produces an image with  96.21\% of the pixels agreeing with the original one, while SA yields 99.19\%.

\section{Discussion}

Both Iterative Conditional Modes (ICM) and Simulated Annealing (SA) effectively restored the majority of the original image; however, the results achieved by SA are significantly superior in visual quality. This difference can be attributed to SA's capability to approximate the global optimum, whereas ICM is often limited to local optima, which can result in subpar restoration outcomes.

Visual analysis of the denoised images indicates that SA effectively eliminated most outlying noise, particularly in uniform regions of the image. However, the reliance on a 4-adjacency neighborhood in the energy function \(E\) restricts SA's ability to capture more complex attributes of the image, such as edges and finer details. The inherent randomness of the SA approach introduces an element of uncertainty, leading to occasional misclassification of pixels, especially in areas where noise obscures the edges of objects. In regions with a high density of noise, certain noise patterns were clustered rather than removed, resulting in residual artifacts.

Despite these challenges, SA's overall performance in restoring the original image structure is commendable. The results highlight the strengths and limitations of both approaches, providing valuable insights into the effectiveness of global optimization techniques in image denoising.

\section{Conclusion}

In this study, we explored the problem of binary image denoising using two distinct methodologies: Iterative Conditional Modes (ICM) and Simulated Annealing (SA). Our experimental results demonstrated that while both methods succeeded in restoring the original image, SA provided a more robust and visually appealing outcome. The ability of SA to approximate the global optimum played a crucial role in its effectiveness, particularly in areas where noise was prevalent.

This research underscores the potential of integrating global optimization techniques, like SA, in the field of image processing. Future work may focus on refining the neighborhood definitions within the energy function and incorporating additional contextual information to enhance edge detection further and preserve fine details. By addressing these limitations, we can improve the overall effectiveness of image-denoising strategies, paving the way for more accurate and reliable applications in various domains, including medical imaging and computer vision.

\end{document}